\documentclass{article} 
\usepackage{style/main,times}

\usepackage{amsmath,amsfonts,bm}

\def\eqref#1{Equation~\ref{#1}}

\def\1{\bm{1}}

\DeclareMathAlphabet{\mathsfit}{\encodingdefault}{\sfdefault}{m}{sl}
\SetMathAlphabet{\mathsfit}{bold}{\encodingdefault}{\sfdefault}{bx}{n}

\usepackage{hyperref}
\usepackage{url}
\usepackage{float}
\usepackage{subcaption}
\usepackage{glossaries-extra}
\setabbreviationstyle[abbreviation]{long-short}

\newabbreviation{vsmax}{VSmax}{Veronese soft-argmax}

\newabbreviation{rht}{RHT}{Reverse mapping of Hough transform}

\newabbreviation{ea}{EA-score}{Euclidean and Angular distance score}

\newabbreviation{id}{ID}{in-distribution}
\newabbreviation{ood}{OOD}{out-of-distribution}

\usepackage{pgfplots}
\usepackage{pgfplotstable}
\usepgfplotslibrary{fillbetween}
\usepgfplotslibrary{groupplots}
\usepackage{booktabs}
\pgfplotsset{compat=1.18}

\definecolor{colBase}{HTML}{7A7A7A}
\definecolor{colOursA}{HTML}{6BAED6}
\definecolor{colOursB}{HTML}{08519C}
\definecolor{colDsnt}{HTML}{E6550D}   
\definecolor{colMlp}{HTML}{31A354}    
\definecolor{colHough}{HTML}{756BB1}  

\pgfplotsset{
  paper/.style={
    width=\linewidth, height=4.6cm,
    tick align=outside, tick pos=left,
    axis line style={gray!60},
    grid=major, grid style={gray!20, line width=0.3pt},
    label style={font=\small},
    tick label style={font=\footnotesize},
    legend cell align=left,
    legend style={font=\footnotesize, draw=none, fill=none},
  },
  base/.style ={colBase,  line width=1.0pt, densely dashed, mark=none},
  oursA/.style={colOursA, line width=1.0pt, mark=none},
  oursB/.style={colOursB, line width=1.1pt, mark=none},
  toplegend/.style={
    legend style={at={(0.5,1.02)}, anchor=south, draw=none, fill=none, font=\footnotesize},
    legend columns=-1,
    /tikz/every even column/.append style={column sep=10pt},
  },
}

\newcommand{\lblvspolartwo}{\gls{vsmax} $+\ \mathcal{L}_{\text{polar}}$}

\newcommand{\lblvsvetwo}{\gls{vsmax} $+\ \mathcal{L}_{\text{vs}}$}

\newcommand{\lbldsntpolar}{Soft-argmax $+\ \mathcal{L}_{\text{polar}}$}
\newcommand{\lbldsntve}{Soft-argmax $+\ \mathcal{L}_{\text{vs}}$}

\newcommand{\lblmlppolar}{MLP $+\ \mathcal{L}_{\text{polar}}$}
\newcommand{\lblmlpve}{MLP $+\ \mathcal{L}_{\text{vs}}$}

\newcommand{\arxivsourcefiles}{%
	\section{Introduction}

Recovering a geometric object from an image requires committing to a parametrisation of it. A line
in the plane admits many: a pair of endpoints, a slope and an intercept, or an orientation and a
signed offset. For a classical algorithm the choice is a matter of convenience. For a neural network
it is decisive, because the network has to \emph{learn} the parametrisation, and what it produces
is a vector in a linear space. The representation must therefore carry the geometry of the object
into a space where linear operations mean something.

Two families of differentiable pipelines regress the parameters of a line (Figure~\ref{fig:pipeline}).
The fully learned pipeline of Figure~\ref{fig:pipeline-a} maps the image to the components of a parametrisation $\vec{p}$ directly
and must discover the geometry of lines from data alone. The known-operator pipeline of
Figure~\ref{fig:pipeline-b} instead inserts a fixed differentiable operator that already encodes that
geometry, which lowers the maximum error bound and the number of free parameters
\citep{known_operators}. For lines that operator is the Hough transform \citep{ht}. It bins every
line by its orientation $\theta$ and signed offset $\rho$, giving $\vec{p}=(\theta,\rho)$.
So a line in the image becomes a peak in the resulting accumulator, and recovering the line reduces to reading out that peak's location.

The Hough accumulator hence parametrises the space of undirected lines, but as a flat rectangle.
However, this space is not flat. 
A line has no direction, so $(\theta,\rho)$ and $(\theta+\pi,-\rho)$ denote the same one.
Per convention the accumulator indexes only the half-circle $\theta\in[0,\pi)$ such that it stays unique.
The two angular edges are therefore glued with a reflection in $\rho$, and the resulting space is a Möbius strip.

The readout is where this matters. A hard argmax selects a single cell and is well defined on the strip, but it passes no gradient. 
Differentiable pipelines use a differentiable version of the argmax, which returns the probability-weighted average of the cell coordinates.
Averaging presupposes a linear space. 
Across the seam it is thus ill posed: two adjacent lines may get averaged into a third that resembles neither, even though the accumulator mass is
concentrated on nearly the same line. Hough-based pipelines either inherit this readout unchanged or
avoid it with a separate, non-differentiable peak extraction.

Two routes lead out. One redefines the operation non-trivially on the curved space. We take the other and change
the representation, so that the linear operation becomes well posed. A line is described by a unit
homogeneous vector $\ell$ that is fixed only up to sign, and we replace it by the outer product
$\ell\ell^{\top}$. Thanks to sign-invariance, both representatives of a line map to the same point, and that
point lives in a linear space. Our proposed \gls{vsmax} operates in this space and returns a barycentre of symmetric
matrices. The line it encodes is recovered in closed form as the leading eigenvector.

\begin{figure}[t]
  \centering
\begin{tikzpicture}[>=stealth, font=\small,
  block/.style={rounded corners=3pt, draw=gray!70, fill=gray!8, font=\footnotesize,
                minimum width=1.9cm, minimum height=0.96cm, align=center},
  ours/.style={rounded corners=3pt, draw=healthy_orange, line width=0.9pt, fill=healthy_orange!10,
               font=\footnotesize, minimum width=1.9cm, minimum height=0.96cm, align=center},
  cnn/.style={rounded corners=2pt, draw=gray!70, fill=gray!8,
              minimum width=0.7cm, minimum height=0.4cm, inner sep=1pt, font=\scriptsize},
  grad/.style={gray!80, opacity=0.4, line width=0.6pt, ->},
  wire/.style={gray!60, line width=0.3pt}]

\begin{scope}[yshift=2.4cm]
  \node[font=\small\bfseries, anchor=east] at (-0.35,0) {(a)};

  \begin{scope}
    \clip (0,-0.75) rectangle (1.5,0.75);
    \fill[black] (0,-0.75) rectangle (1.5,0.75);
    \draw[white, line width=0.9pt] (0.15,-0.9) -- (1.35,0.9);
  \end{scope}
  \draw[gray!60] (0,-0.75) rectangle (1.5,0.75);

  \draw[->]   (1.55,-0.10) -- (2.00,-0.10);
  \draw[grad] (2.00, 0.10) -- (1.55, 0.10);

  \node[cnn, minimum width=1.0cm, minimum height=0.9cm] at (2.55,0) {CNN};

  \foreach \y in {0.45,0.15,-0.15,-0.45}{%
    \foreach \z in {0.2,-0.2}{\draw[wire] (3.425,\y) -- (4.425,\z);}}
  \foreach \y in {0.45,0.15,-0.15,-0.45}{\fill[gray!55] (3.425,\y) circle (1.6pt);}
  \foreach \z in {0.2,-0.2}{\fill[gray!55] (4.425,\z) circle (1.6pt);}
  \node[anchor=west, font=\footnotesize] at (4.575,0.2) {$p_1$};
  \node[anchor=west, font=\footnotesize] at (4.575,-0.2) {$p_2$};
  \draw[gray!70, line width=0.5pt, decorate, decoration={brace, amplitude=4pt}]
    (5.025,0.33) -- (5.025,-0.33) node[midway, right=5pt, font=\footnotesize, text=black]{$\vec{p}$};

\end{scope}

\begin{scope}
  \node[font=\small\bfseries, anchor=east] at (-0.35,0) {(b)};

  \begin{scope}
    \clip (0,-0.75) rectangle (1.5,0.75);
    \fill[black] (0,-0.75) rectangle (1.5,0.75);
    \draw[white, line width=0.9pt] (0.15,-0.9) -- (1.35,0.9);
  \end{scope}
  \draw[gray!60] (0,-0.75) rectangle (1.5,0.75);

  \draw[->]   (1.55,-0.10) -- (2.00,-0.10);
  \draw[grad] (2.00, 0.10) -- (1.55, 0.10);

  \node[block] at (3.00,0) {HT-Layer};

  \draw[->]   (4.00,-0.10) -- (4.45,-0.10);
  \draw[grad] (4.45, 0.10) -- (4.00, 0.10);

  \begin{scope}[shift={(4.50,-0.48)}]
    \begin{scope}
      \clip (0,0) rectangle (0.96,0.96);
      \fill[black] (0,0) rectangle (0.96,0.96);
      \foreach \s in {-0.8,-0.7,...,0.8}{%
        \draw[white, opacity=0.3, line width=0.2pt]
          plot[domain=0:180, samples=80, smooth]
            ({0.96*\x/180}, {0.48+0.48*(0.2*cos(\x-115)+\s*sin(\x-115))});}
      \fill[white] (0.613,0.576) circle (1.3pt);
    \end{scope}
    \draw[gray!60] (0,0) rectangle (0.96,0.96);
  \end{scope}

  \draw[->]   (5.51,-0.10) -- (5.96,-0.10);
  \draw[grad] (5.96, 0.10) -- (5.51, 0.10);

  \node[cnn] at (6.36,0) {CNN};

  \draw[->]   (6.76,-0.10) -- (7.21,-0.10);
  \draw[grad] (7.21, 0.10) -- (6.76, 0.10);

  \begin{scope}[shift={(7.26,-0.48)}]
    \fill[black] (0,0) rectangle (0.96,0.96);
    \fill[white] (0.613,0.576) circle (1.3pt);
    \draw[gray!60] (0,0) rectangle (0.96,0.96);
  \end{scope}

  \draw[->]   (8.27,-0.10) -- (8.72,-0.10);
  \draw[grad] (8.72, 0.10) -- (8.27, 0.10);

  \node[ours] at (9.72,0) {Readout};

  \draw[->]   (10.72,-0.10) -- (11.17,-0.10);
  \draw[grad] (11.17, 0.10) -- (10.72, 0.10);
  \node[anchor=west] at (11.22,0) {$\vec{p}$};
\end{scope}

\end{tikzpicture}%
  \begin{subfigure}{0.0001\linewidth}\phantomsubcaption\label{fig:pipeline-a}\end{subfigure}%
  \begin{subfigure}{0.0001\linewidth}\phantomsubcaption\label{fig:pipeline-b}\end{subfigure}%
  \caption{Two differentiable pipelines that regress the parameters of a line. (a) A fully learned pipeline regresses a parametrisation $\vec{p}$ and
  must discover line geometry from data. (b) A known-operator pipeline applies a fixed Hough
  transform and extracts the peak with a readout. We build on (b) and study the readout.}
  \label{fig:pipeline}
\end{figure}
	\section{Related Work}

\paragraph{Known operators and the Hough transform.}
Known-operator learning replaces parts of a network by fixed differentiable operators that encode
prior knowledge; \citet{known_operators} prove this lowers both the maximum error bound and the
number of free parameters. For lines the natural such operator is the Hough transform, introduced
in its polar form by \citet{ht}. \citet{dht} embed the transform in a CNN but read out lines with a
separate, non-differentiable peak-extraction step, so gradients never flow through the readout.

\paragraph{Differentiable coordinate readouts.}
\citet{softargmax} and \citet{dsnt} make peak extraction differentiable the same way.
Normalise a heatmap into a distribution over cells, then return the probability-weighted average of a cached
coordinate grid. They differ only in how that distribution is formed, not in what they return: 
a point in the convex hull of the grid. \citet{dsnt} write the average as an inner product with
fixed coordinate matrices, improving on heatmap matching, which is not differentiable up to the
coordinate, and on fully connected regression, which lacks spatial generalisation.
Every such readout presumes a chart whose coordinates may be averaged linearly. \citet{dsat} come
closest to questioning this, wrapping the soft-argmax circularly in $\theta$; but auto-rotation
needs no offset, so they discard $\rho$ and never meet the coupling
$(\theta,\rho)\sim(\theta+\pi,-\rho)$ that makes such an average ill posed.

\paragraph{Symmetry-aware architectures.}
Where a domain carries a symmetry, the standard remedy is equivariant layers. \citet{group_conv}
generalise convolution to groups generated by translations, reflections and rotations. At the
angular boundary of the Hough space the identification is a glide-reflection
\citep{glide_reflection}, an orientation-reversing isometry to which a translation-only convolution
is not equivariant. Unfolding the angular axis from $[0,\pi)$ to the full circle $[0,2\pi)$ replaces
it by an ordinary circular wrap, so the accumulator becomes a cylinder and the layer a
\emph{cylindrical convolution} \citep{cylinder_convolution}, for which circular padding is exact and
plain translation equivariance suffices. The unfolded accumulator, however, covers every undirected
line twice, so no standard readout is invariant to both representatives.

\paragraph{Projection embeddings.}
An undirected line is a vector fixed only up to sign, with no linear structure of its own. Several
fields resolve this by embedding such an object as the outer product of its vector, a symmetric matrix
on which Euclidean computations apply. \citet{polynomial_kernel} build this into the projection kernel and
state the mismatch plainly: feature extraction carried out in a Euclidean space while the distances
that matter are not Euclidean. The same embedding underlies kernels that inherit the ambient metric
\citep{projective_embedding}, dictionary learning in Frobenius norm \citep{dictionary_learning}, and
network layers returning Euclidean forms to ordinary output layers \citep{outter_product}. Recovering
the vector from the matrix is a low-rank approximation, solved in closed form by
\citet{eckart_young}. Outside learning, \citet{chordal_distance_transform} reach the same Möbius band
and chordal distance once pairs of points on a circle are identified. All of them run the embedding
forward, from data to a feature they describe or classify; none runs it backward as a readout,
turning a distribution over lines into a single recovered line.

\paragraph{Lifted representations and geometric losses.}
3D Rotation estimation meets the same problem and answers it by lifting: represent a rotation in a
higher-dimensional linear space, regress there, and project back. \citet{9d_representation} predict
nine numbers and project onto $SO(3)$ by SVD, which outperforms quaternions.
\citet{quaternion_averaging} average quaternions, which carry the same sign ambiguity as an
undirected line, by maximising a quadratic form, precisely the recovery such an embedding calls for.
\citet{rotation_averaging} define the chordal distance between rotations as the Euclidean distance
between their $\mathbb{R}^9$ embeddings, and \citet{geodesic_loss} confirm that such a geometric loss
beats a plain Euclidean one, though their geodesic distance on $SO(3)$ has a gradient that explodes
when two rotations are half a turn apart. So the lift, the recovery,
and the geometric loss are already in place, but only for a group: $SO(3)$ is compact with a
bi-invariant metric and an orientable double cover, whereas undirected lines form a non-orientable
quotient that is no group at all.

\paragraph{Positioning.}
Each field holds part of the answer, but none is built for lines: vision contributes the
known-operator pipeline and a differentiable readout yet treats the accumulator as a flat image;
subspace learning contributes a sign-invariant representation and its inverse yet uses both only to
describe objects it already has; and rotation estimation contributes lift, recovery and geometric
loss yet for a fundamentally different space. What none provides is a differentiable readout well
posed on the space of undirected lines, together with an objective that measures genuine line distance.%
	\section{Methods}

\subsection{A space-operation mismatch}
\label{space-operation_chapter}

An undirected line is described in Hough coordinates by an orientation
$\theta\in[0,\pi)$ and a signed offset $\rho\in\mathbb{R}$, and the Hough
transform indexes its accumulator by exactly these polar coordinates $(\theta,\rho)$.
The half-circle therefore provides one representative for every undirected
line. However, the underlying representation is still subject
to the identification:

\begin{equation}
    \label{eq:seam-identification}
  (\theta,\rho)\sim(\theta+\pi,-\rho).
\end{equation}

Identifying the angular boundaries according to this relation yields the topology of a Möbius strip.
Crossing from $\theta=\pi$ back to $\theta=0$ flips the sign of $\rho$ (Figure~\ref{fig:concept}).
Consequently, two geometrically adjacent lines can lie at opposite ends of the
finite chart. This is harmless for a non-differentiable readout: a hard argmax simply selects the
peak cell, a well-defined operation on the strip. Training a neural network, however, requires a
differentiable readout, the soft-argmax, which returns a weighted average of the cell coordinates.
Averaging presupposes a linear space, and the strip is not one. Across the seam the soft-argmax is
therefore ill posed: it can average two geometrically adjacent lines into one far from both, even though the
accumulator mass is concentrated on nearly identical lines. 
This seam is an artefact of the 2D coordinate system, not of the line space itself.
Undirected lines form a smooth manifold, and any flat chart of it must cut somewhere.
The next section makes this space precise and derives a representation in which a readout can be defined that is well posed.

\begin{figure}[t]
  \centering
\begin{tikzpicture}[>=stealth, font=\footnotesize]

\begin{scope}
  \draw[gray!60] (0,0) rectangle (3.4,3.4);
  \draw[healthy_orange, line width=1pt] (2.30,0) -- (2.50,3.4);
  \draw[siemens_petrol, line width=1pt, densely dashed] (2.50,0) -- (2.30,3.4);
  \node[healthy_orange, anchor=west] at (2.60,0.80) {$\ell_1$};
  \node[siemens_petrol,  anchor=east] at (2.20,2.60) {$\ell_2$};
  \node[align=center] at (1.7,-0.95) {image space\\[-2pt]\scriptsize two nearly identical lines};
\end{scope}

\begin{scope}[xshift=5.2cm]
  \draw[gray!60] (0,0) rectangle (3.4,3.4);
  \draw[gray!35, dashed] (0,1.7) -- (3.4,1.7);

  \fill[healthy_orange] (0.16,2.66) circle (2pt);
  \node[healthy_orange, anchor=west] at (0.30,2.66) {$\ell_1$};
  \fill[siemens_petrol] (3.24,0.74) circle (2pt);
  \node[siemens_petrol, anchor=east] at (3.10,0.74) {$\ell_2$};

  \draw[healthy_orange, ->, line width=1.4pt] (0,0.35) -- (0,3.05);
  \draw[siemens_petrol,  ->, line width=1.4pt] (3.4,3.05) -- (3.4,0.35);

  \node[anchor=north] at (0,-0.05) {\scriptsize $0$};
  \node[anchor=north] at (3.4,-0.05) {\scriptsize $\pi$};
  \node[anchor=north] at (1.7,-0.05) {$\theta$};
  \node[rotate=90] at (-0.38,1.7) {$\rho$};
  \node[align=center] at (1.7,-0.95) {Hough space\\[-2pt]\scriptsize opposite corners, glued with a flip};
\end{scope}

\end{tikzpicture}
  \caption{Two nearly identical lines $\ell_1,\ell_2$ (left) map to opposite ends of the Hough
  space (right); the $\theta=0$ and $\theta=\pi$ edges are glued with a sign flip on $\rho$, so a
  weighted average across the seam is ill posed.}
  \label{fig:concept}
\end{figure}

\subsection{The Veronese embedding}
\label{sec:veronese}

In Hesse normal form, a line satisfies $\rho=x\cos\theta+y\sin\theta$.
Its coefficient vector $(\cos\theta,\ \sin\theta,\ -\rho)^{\top}$ describes the line only up to nonzero scaling $\lambda\in\mathbb{R}\setminus\{0\}$.
Through normalisation to unit length, the scaling reduces to a sign ambiguity $\lambda=\pm1$, and a line $\ell$ can be described by

\begin{equation}
  \label{eq:line-parametrization}
  \ell \;=\; \frac{1}{\sqrt{1+\rho^2}}\,(\cos\theta,\ \sin\theta,\ -\rho)^{\top}\ \in\ S^2\subset\mathbb{R}^3,
\end{equation}

where $S^2$ is the unit sphere in $\mathbb{R}^3$. 
On $S^2$ the polar seam identification in \eqref{eq:seam-identification} becomes $\ell\sim-\ell$.
We seek a space where $\ell$ and $-\ell$ are represented by the same point.
This translates to gluing each point on $S^2$ to its antipode, which is algebraically formalised as the quotient space $S^2/\{\pm1\}$.
The quotient is by definition the real projective plane $\mathbb{RP}^2$.
The Veronese map $\nu_d$ \citep{veronese_map} sends a projective space $\mathbb{P}^n$ into a higher-dimensional projective space $\mathbb{P}^{N}$ by representing a point $[x_0:\dots:x_n]$ through all its degree-$d$ monomials,

\begin{equation}
  \nu_d\colon\ \mathbb{P}^n \to \mathbb{P}^N,\qquad
  [x_0:\dots:x_n]\ \longmapsto\ \bigl[\,x_0^{d}:x_0^{d-1}x_1:\dots:x_n^{d}\,\bigr].
\end{equation}

where $d$ is the degree, $n$ the dimension of the source space, and $N=\binom{n+d}{d}-1$ the dimension of the target. 
Taking the monomials as ordinary coordinates places $\mathbb{P}^n$ inside the linear space $\mathbb{R}^{N+1}$.
In our case, the source dimension is fixed by the projective plane, so $n=2$.
Moreover, the $d=2$ monomials $x_ix_j$ are exactly the entries of the outer product $\ell\ell^{\top}$, a rank-one symmetric matrix.
Two steps turn the projective map $\nu_2$ into a concrete embedding we can work with. 
First, we exchange the projective class for our definite unit-norm representative $\ell\in S^2$.
The quadratic monomials $\ell_i\ell_j$ now absorb the sign ambiguity, since $\ell_i\ell_j=(-\ell_i)(-\ell_j)$.
Second, since $\ell\ell^{\top}$ is symmetric we keep only its six distinct entries, obtained by half-vectorisation $\operatorname{vech}(\ell\ell^{\top})$.
The result is the Veronese embedding as a vector-space realisation $v_2(\ell):S^2\to\mathbb{R}^6$, defined as:

\begin{equation}
  v_2(\ell) \;=\; \operatorname{vech}(\ell\ell^{\top}) \;=\; \bigl(\ell_1^2,\ \ell_2^2,\ \ell_3^2,\ \ell_1\ell_2,\ \ell_1\ell_3,\ \ell_2\ell_3\bigr)^{\top},\qquad v_2(\ell)=v_2(-\ell).
\end{equation}

This embeds $\mathbb{RP}^2$ into the \emph{linear} space $\mathbb{R}^6$: the seam vanishes, antipodal cells coincide, and geometric proximity of lines becomes metric proximity.
In particular, the representation inherits the Euclidean metric ($\lVert\cdot\rVert_2$ for vectors, $\lVert\cdot\rVert_F$ for matrices) of the ambient space.
So a standard $L_2$ loss on $v_2$ already is a valid training objective.
For that loss to measure the true error between lines $\ell$, the embedding must become isometric.
In its plain form each off-diagonal entry appears once in $\operatorname{vech}(\ell\ell^{\top})$ but twice in $\ell\ell^{\top}$, so we apply the diagonal weighting matrix $W=\operatorname{diag}(1,1,1,\sqrt2,\sqrt2,\sqrt2)$:

\begin{equation}
  \lVert W v_2(\ell)\rVert_2 \;=\; \lVert \ell\ell^{\top}\rVert_F \;=\; \lVert\ell\rVert_2^2.
\end{equation}

We fold this weighting into $v_2$ hereafter.
The squared $L_2$ distance of weighted Veronese embeddings now equals the squared chordal distance \citep{rotation_averaging} between lines in projective space (Appendix~\ref{app:proof}).

A \emph{standard} objective function used in deep learning acquires an exact projective-geometric meaning.
This is the strategy we follow throughout. 
Take the convolutions that produce the accumulator weights in Figure~\ref{fig:pipeline-b}.
Any kernel with a receptive field wider than a single cell must pad beyond the grid boundary, 
where the identification in \eqref{eq:seam-identification} acts as a \emph{glide-reflection} \citep{glide_reflection} in $\rho$ (see Section~\ref{space-operation_chapter}). 
Standard convolutions are equivariant to translations but not to reflections.    
Since antipodal partners coincide in our representation, we can spatially unfold the accumulator onto the double cover $[0,2\pi)$.
Both covers together carry an ambiguous representation of the same undirected line, 
but the convolution's role is to \emph{clean} the accumulator grid:
a task invariant to which antipodal representative a cell carries.
On the double cover the seam is no longer a glide-reflection
but an ordinary circular wrap, so the accumulator becomes a cylinder and the layer reduces to a
\emph{cylindrical convolution} \citep{cylinder_convolution}, for which plain circular padding is
exact and topologically correct.

\subsection{Veronese soft-argmax}

Given our representation, the intrinsics of the CNN and the objective function are set, the missing piece is a readout being well defined on the projective plane.
As described in Section~\ref{space-operation_chapter}, the standard soft-argmax is not.
We seek to remove its bottleneck with the \emph{\gls{vsmax}}, a readout that becomes well defined in the projective plane.
The idea is to apply the soft-argmax not on the 2D accumulator but on the 6D Veronese embedding.
The Hough transform bins lines into a fixed grid of $N$ accumulator cells, so we cache one column per cell as a correspondence tensor.
For every cell $(\theta,\rho)$ soft-argmax caches the bare polar coordinates as $C\in\mathbb{R}^{2\times N}$, while \gls{vsmax} caches the Veronese embedding as $V\in\mathbb{R}^{6\times N}$:

\begin{equation}
  \underbrace{C \;=\; \bigl[\,(\theta,\rho)\,\bigr]_{(\theta,\rho)} \ \in\ \mathbb{R}^{2\times N}}_{\text{Polar Correspondence Tensor}}
  \qquad\longrightarrow\qquad
  \underbrace{V \;=\; \bigl[\,v_2(\ell(\theta,\rho))\,\bigr]_{(\theta,\rho)} \ \in\ \mathbb{R}^{6\times N}}_{\text{Veronese Correspondence Tensor}}.
\end{equation}

Both operators average the same softmax-activated probability distribution $P$ over the grid, ensuring that

\begin{equation}
    P((\theta,\rho))\ge0, \quad \sum_{\theta,\rho} P((\theta,\rho))=1.
\end{equation}

Full vectorisation of this grid, $p=\operatorname{vec}(P)\in\mathbb{R}^N$, lets each readout be written as a single matrix--vector product with its cached tensor.
Replacing the averaged coordinates with their embeddings defines the \gls{vsmax}:

\begin{equation}
  \underbrace{(\hat\theta,\hat\rho)=\sum_{\theta,\rho} P((\theta,\rho))\,(\theta,\rho)= C\,p}_{\text{soft-argmax}}
  \qquad\longrightarrow\qquad
  \underbrace{\hat v=\sum_{\theta,\rho} P(\theta,\rho)\, v_2(\ell(\theta,\rho)) = V\,p}_{\text{VSmax}}.
\end{equation}

The two operators utilise the same operation in different spaces, hence, they differ in their output.
Soft-argmax returns directly Hough-native polar coordinates, but averaged in the torn coordinate system.
\gls{vsmax} instead returns the barycentre of the selected lines in the embedding: seam-free and linear, but not yet a line $\pm\ell$.
Hence, we need a decoding mechanism mapping the estimate $\hat v$ to polar coordinates.

\subsection{Line recovery}

The \gls{vsmax} returns a barycentre $\hat v$, a point in the Veronese embedding instead of a line.
Recovering the one it encodes takes three differentiable steps.
First, inverting the isometric weighting $W$ and the half-vectorisation folds $\hat v$ back into the symmetric matrix $\hat V$ it encodes,

\begin{equation}
  \hat V \;=\; \operatorname{vech}^{-1}\!\bigl(W^{-1}\hat v\bigr) \;\in\; \mathrm{Sym}^2(\mathbb{R}^3).
\end{equation}

In Section~\ref{sec:veronese} we saw that a single line embeds as the sign-invariant rank-one projector onto the line's direction. 
The barycentre $\hat V$, however, is a convex combination of such projectors and is in general no longer rank-one.
The second step therefore is to find the rank-one symmetric matrix closest to $\hat V$.
According to the Eckart--Young theorem \citep{eckart_young}, this is solved by the following maximisation problem

\begin{equation}
  \pm\hat\ell \;=\; \arg\max_{\lVert u\rVert_2=1} u^\top \hat V\, u ,
\end{equation}

whose solution is the leading eigenvector of $\hat V$, recovered in closed form.
Finally, $\hat\ell=(\hat\ell_1,\hat\ell_2,\hat\ell_3)$ is the unit homogeneous vector of \eqref{eq:line-parametrization}, so inverting that parametrisation returns the line to Hough coordinates.
The third and last step is therefore:

\begin{equation}
  \hat\theta \;=\; \arctan\!\left(\frac{\hat\ell_2}{\hat\ell_1}\right)
             \;=\; \arctan\!\left(\frac{\sin\theta}{\cos\theta}\right),
  \qquad
  \hat\rho \;=\; -\,\frac{\hat\ell_3}{\sqrt{\hat\ell_1^2+\hat\ell_2^2}}
           \;=\; -\bigl(-\rho\bigr) .
\end{equation}

Regarding end-to-end differentiation, only the eigenvalue decomposition is delicate.
It is differentiable wherever the largest eigenvalue is strictly greater than the second, and its gradient is ill conditioned only as that gap closes.
This matters where the recovery itself must be a robust end-to-end differentiable component. For our proposed method, it does not affect us. 
The loss acts directly on the embedding $\hat v$, leaving the eigenvector extraction and $(\hat\theta,\hat\rho)$ decoding as post-processing outside the gradient path.%
	\section{Experiments and Results}

We study line recovery through the known-operator pipeline in Figure~\ref{fig:pipeline}(b): a fixed
Hough transform maps the image to the $(\theta,\rho)$ accumulator, a small refinement network cleans
it, and a differentiable \emph{readout} extracts the line parametrisation. 
Two modular blocks vary across our main experiments: the \emph{readout} (soft-argmax vs.\ \gls{vsmax}), 
and the coordinate \emph{representation} in which the $L_2$ loss is taken (native polar $(\theta,\rho)$ vs.\ the Veronese embedding):

\begin{equation}
  \mathcal{L}_{\text{polar}} = \bigl\lVert(\theta^\star,\rho^\star)-(\hat\theta,\hat\rho)\bigr\rVert_2^2,
  \qquad
  \mathcal{L}_{\text{vs}} = \bigl\lVert v^\star - \hat v\bigr\rVert_2^2,
\end{equation}

Convolutions on the single-cover Hough space (all non-\gls{vsmax} variants) pad with the strategy defined in Appendix~\ref{app:mobius}, 
the most faithful single-cover choice.
For \gls{vsmax} variants we unfold the accumulator onto the double cover, where ordinary circular padding is exact.
Any other hyperparameters or network configurations are fixed across experiments and can be found in Appendix~\ref{app:setup}.
In the ablations, we additionally introduce three baselines that each break one ingredient.
A data-driven backbone CNN (ResNet-18) with a multi-layer perceptron (MLP) readout, a learned Hough-heatmap matching approach with a non-differentiable readout, and the classical Hough transform with no learning at all.

We train on synthetic $256\times256$ images, each showing a single line at a random
$(\theta,\rho)$ drawn uniformly over all resolvable Hough space lines at $127\times127$ resolution ($N=14384$), with additive Gaussian pixel noise of
standard deviation $\sigma$ as the difficulty knob and a fixed seed for comparability. 
Training draws $\sigma$ uniformly from $[0,0.5]$, and a held-out validation split ($N=1000$) from the same range is used for model
selection.
We then test on $\sigma\in\{0,\dots,0.8\}$, so that $\sigma>0.5$ probes \gls{ood}
robustness beyond the training range. All Hough-based configurations share the fixed Hough transform,
the same $37.6$K-parameter refinement network, and the same optimiser, schedule and seed, so that only
the readout and the representation vary (see Appendix~\ref{app:setup}).

For a predicted line $\hat{\ell}$ and ground-truth line $\ell^\star$, we use the \gls{ea} of
\citet{dht} as the evaluation metric, which combines angular ($S_\theta$) and positional ($S_d$) similarity. Let $\Delta\theta\in[0,\pi/2]$ be the
angle between the two lines and $d_{\ell}$ the Euclidean distance between the midpoints of their
visible intersections with the image domain, after normalising the image to a unit square. With

\begin{equation}
  S_\theta = 1-\frac{\Delta\theta}{\pi/2},
  \qquad
  S_d = 1-{d_{\ell}},
  \qquad
  \operatorname{EA}(\hat{\ell},\ell^\star)
  = (S_\theta S_d)^2 \in [0,1],
\end{equation}

larger values indicate better agreement. 
We report the mean per-line \gls{ea} over all test lines.
Additionally we separate between those scores over the \emph{seam}, the orientation-wrap bins $\theta$-bin $\in\{0,\dots,4\}\cup\{122,\dots,126\}$,
and over the \emph{interior}, the remaining bins.

\subsection{Operator validation}

Fixing $\mathcal{L}_{\text{vs}}$, we change only the readout and test \gls{ood} performance at $\sigma=0.6$.
Figure~\ref{fig:boundary} colour-codes the \gls{ea} of every resolvable line bin over the full
$(\theta,\rho)$ accumulator. Using soft-argmax is relatively accurate in the interior but
tears at the orientation-wrap seam. Specifically, seam \gls{ea} $0.69$ against $0.72$ in the interior (Table~\ref{tab:main}) underlines the failure of the space-operation mismatch.
We show that using \gls{vsmax} resolves it: antipodal cells share an embedding point, so mass split across the seam reunites, and seam
\gls{ea} rises to $0.92$, matching its interior ($0.88$).
Both readouts nonetheless leave a band of outliers spanning the full angular range at the extreme
offsets, where $\rho$ is near its maximum or minimum. 
These are expected, and not a topological defect.
Those parametrisations correspond to lines that are far from the origin (e.g.\ at image corners) leaving only a small segment of the line visible in the image.
The Hough response is weak, more so under noise.
Even in this regime, \gls{vsmax} shrinks the outlier bands relative to soft-argmax.

\begin{figure}[t]
  \centering
\begin{tikzpicture}
\begin{groupplot}[
  group style={group size=2 by 1, horizontal sep=0.9cm},
  width=3.2cm, height=3.2cm, scale only axis,
  enlargelimits=false, axis on top,
  xmin=0, xmax=127, ymin=0, ymax=127,
  xtick={0,63,126}, ytick={0,63,126},
  tick label style={font=\scriptsize}, title style={font=\footnotesize},
  xlabel={$\theta$ bin}, xlabel style={font=\scriptsize},
  colormap/viridis, point meta min=0.4, point meta max=0.95,
]
\nextgroupplot[title={soft-argmax}, ylabel={$\rho$ bin}, ylabel style={font=\scriptsize}]
  \addplot graphics[xmin=0, xmax=127, ymin=0, ymax=127]{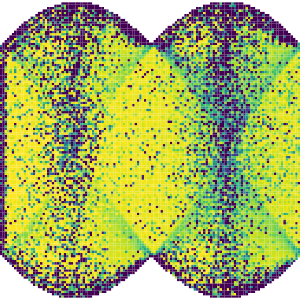};
\nextgroupplot[title={VSmax},
    colorbar, colorbar style={width=5pt, ylabel={EA-score}, ylabel style={font=\scriptsize},
    ytick={0.4,0.6,0.8}, tick label style={font=\scriptsize}}]
  \addplot graphics[xmin=0, xmax=127, ymin=0, ymax=127]{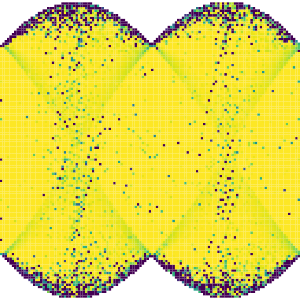};
\end{groupplot}
\end{tikzpicture}
  \caption{\glsxtrshortpl{ea} of every resolvable and visible line, colour-coded over the full $(\theta,\rho)$ Hough accumulator at
  $\sigma=0.6$ (\gls{ood}), with the Veronese representation fixed and only the readout
  changed. Standard soft-argmax (left) is accurate in the interior but dark at the orientation-wrap
  seam ($\theta$-bin $0/126$); the Veronese soft-argmax (right) leaves a near-uniform map.}
  \label{fig:boundary}
\end{figure}

\subsection{Representation validation}

We now follow the inverse approach. 
Fixing each readout in turn, we exchange only the representation in which the $L_2$-loss is measured: $\mathcal{L}_{\text{polar}}$ vs.\ $\mathcal{L}_{\text{vs}}$ (Table~\ref{tab:main}).
The Veronese representation lifts the seam \gls{ea} in \emph{every} configuration. 
The clearest case is the MLP, which has no readout to blame: switching its loss to the embedding
lifts seam \gls{ea} by $28\%$, closing the seam deficit so that it matches the interior. The same swap
lifts standard soft-argmax by $10\%$ and our \gls{vsmax} by $14\%$.

Two things follow. 
First, $\mathcal{L}_{\text{vs}}$ alone is not enough: soft-argmax with the embedding loss still reaches only $0.69$ at the seam, far below \gls{vsmax} ($0.92$).
The objective function constrains the output space, not the operations that produce it.
Hence, representation and operator are complementary, both must respect the projective geometry of lines.
Second, $\mathcal{L}_{\text{polar}}$ and $\mathcal{L}_{\text{vs}}$ trade off in the \emph{interior} as noise grows (Figure~\ref{fig:noise}).
Here the seam is isolated, so the difference is one of metric fidelity.
Minimising $\mathcal{L}_{\text{vs}}$ equals the chordal distance between lines (Section~\ref{sec:veronese}), so it minimises true geometric line error which is exactly what the \gls{ea} rewards.
In contrast, $\mathcal{L}_{\text{polar}}$ penalises raw $(\theta,\rho)$ coordinate error, which does not match line distance.
Since $\ell$ depends nonlinearly on $\rho$, a fixed $(\theta,\rho)$ error corresponds to a large geometric change for some lines, and a tiny one for others.
This results in a distorted measure of line error, because $\mathcal{L}_{\text{polar}}$ weights them all equally.
On clean data this mismatch is negligible and the Hough-native coordinates are beneficial.
As noise increases, $\mathcal{L}_{\text{polar}}$ increasingly optimises this distorted proxy, so its interior
\gls{ea} decreases faster, while $\mathcal{L}_{\text{vs}}$ stays aligned with true line distance and degrades gracefully.

\begin{table}[t]
  \centering
  \caption{\glsxtrshort{ea} at $\sigma=0.6$ (\gls{ood}) per method: global, at the orientation-wrap
  seam, and in the interior. Blocks (top to bottom): data-driven MLP, standard soft-argmax, and our
  Veronese soft-argmax readout, each with the polar and the Veronese representation. Switching to
  the Veronese representation lifts the seam \glsxtrshort{ea} within every pipeline.}
  \label{tab:main}
\newcommand{\runlabelmain}[1]{%
  \ifcase#1
    \lblmlppolar\or    
    \lblmlpve\or       
    \lbldsntpolar\or   
    \lbldsntve\or      
    \lblvspolartwo\or  
    \lblvsvetwo\fi     
}

\pgfplotstabletypeset[
  col sep=space,
  every head row/.style={before row=\toprule, after row=\midrule},
  every last row/.style={after row=\bottomrule},
  every row no. 2/.style={before row=\midrule},
  every row no. 4/.style={before row=\midrule},
  columns={run,ea_all,ea_seam,ea_interior},
  columns/run/.style={
    column name={Method}, column type={l},
    assign cell content/.code={\pgfkeyssetvalue{/pgfplots/table/@cell content}{\runlabelmain{##1}}},
  },
  columns/ea_all/.style={column name={EA (all)},           string type, column type={c}},
  columns/ea_seam/.style={column name={EA (seam)},         string type, column type={c}},
  columns/ea_interior/.style={column name={EA (interior)}, string type, column type={c}},
]{figures/dat/ea_methods_main.dat}

\end{table}

\begin{figure}[h]
  \centering
\begin{tikzpicture}
\begin{groupplot}[
  group style={group size=3 by 1, horizontal sep=1.2cm},
  paper, width=0.30\linewidth, height=4.4cm,
  xmin=0, xmax=0.8, ymin=0, ymax=1.02,
  xtick={0,0.2,0.4,0.6,0.8},
  xlabel={test noise $\sigma$},
  title style={font=\footnotesize},
]
\nextgroupplot[title={MLP}, ylabel={interior EA-score}]
  \addplot[healthy_orange, line width=1.1pt, mark=*, mark size=1pt] table[x=sigma, y=mlpve]{figures/dat/ea_noise_interior.dat};
  \addplot[siemens_petrol, line width=1.0pt, densely dashed, mark=triangle*, mark size=1.3pt] table[x=sigma, y=mlppolar]{figures/dat/ea_noise_interior.dat};
\nextgroupplot[title={soft-argmax}]
  \addplot[healthy_orange, line width=1.1pt, mark=*, mark size=1pt] table[x=sigma, y=dsntve]{figures/dat/ea_noise_interior.dat};
  \addplot[siemens_petrol, line width=1.0pt, densely dashed, mark=triangle*, mark size=1.3pt] table[x=sigma, y=dsntpolar]{figures/dat/ea_noise_interior.dat};
\nextgroupplot[title={VSmax},
  legend pos=south west, legend style={draw=none, fill=none, font=\scriptsize}]
  \addplot[healthy_orange, line width=1.1pt, mark=*, mark size=1pt] table[x=sigma, y=vsve2]{figures/dat/ea_noise_interior.dat};
  \addlegendentry{$\mathcal{L}_{\text{vs}}$}
  \addplot[siemens_petrol, line width=1.0pt, densely dashed, mark=triangle*, mark size=1.3pt] table[x=sigma, y=vspolar2]{figures/dat/ea_noise_interior.dat};
  \addlegendentry{$\mathcal{L}_{\text{polar}}$}
\end{groupplot}
\end{tikzpicture}
  \caption{Interior \glsxtrshort{ea} vs.\ test noise $\sigma$ (dashed line: training maximum $\sigma=0.5$) for the
  three pipelines---MLP, standard soft-argmax, and \glsxtrshort{vsmax}---each with the polar (teal, dashed) and
  the Veronese (orange) representation. The polar loss can lead when clean, but the Veronese loss
  degrades more gracefully as noise grows in every pipeline.}
  \label{fig:noise}
\end{figure}

\subsection{Ablation 1: why learn end-to-end}

We want to show why end-to-end training is beneficial in this setup.
For that, we compare our method to two Hough-variants that do not rely on it.
The plain Hough transform (no refinement), read out by non-differentiable (argmax) peak selection,
and the \gls{rht}~\citep{dht}, a learned heatmap-matching baseline (with refinement) from the multi-line detection literature
with a non-differentiable readout. 
Figure~\ref{fig:endtoend} sweeps the test noise. 
The classical transform is exact when clean ($\approx0.98$) but decays towards $0.5$ as noise grows.
This is exactly the regime a learned network fixes, by cleaning the accumulator before the readout.
The RHT does not transfer to single-line regression: 
at every hyperparameter configuration we tried it collapses under mild noise (we sweep these RHT-specifics in Appendix~\ref{app:rht}).
Its recipe is built to detect an unknown number of lines rather than regress a single one end-to-end.
Only our differentiable \gls{vsmax} pipeline degrades gracefully and stays robust across the sweep.

\begin{figure}[H]
  \centering
\begin{tikzpicture}
\begin{axis}[paper, width=\linewidth, height=5cm,
  xlabel={test noise $\sigma$},
  ylabel={EA-score},
  xmin=0, xmax=0.8, ymin=0, ymax=1.02,
  xtick={0,0.1,0.2,0.3,0.4,0.5,0.6,0.7,0.8},
  toplegend,
]
  \addplot[healthy_orange!55, line width=1.0pt, mark=square*, mark size=1pt] table[x=sigma, y=rht]{figures/dat/ea_noise_full.dat};
  \addlegendentry{RHT}
  \addplot[healthy_orange, line width=1.0pt, densely dashed, mark=diamond*, mark size=1.3pt] table[x=sigma, y=classicalhough]{figures/dat/ea_noise_full.dat};
  \addlegendentry{HT+argmax}
  \addplot[healthy_orange!70!black, line width=1.2pt, mark=*, mark size=1pt] table[x=sigma, y=vsve2]{figures/dat/ea_noise_full.dat};
  \addlegendentry{VSmax}
\end{axis}
\end{tikzpicture}
  \caption{Global \glsxtrshort{ea} on the full test set vs.\ test noise. The classical Hough readout is exact when
  clean but brittle; the RHT (best target width; Appendix~\ref{app:rht})
  collapses under noise; our end-to-end differentiable Veronese soft-argmax degrades gracefully.}
  \label{fig:endtoend}
\end{figure}

\subsection{Ablation 2: why known operators}
\label{sec:generalization}

\begin{figure}[b!]
  \centering
\begin{tikzpicture}
\begin{groupplot}[
  group style={group size=4 by 3, horizontal sep=0.45cm, vertical sep=0.6cm,
    x descriptions at=edge bottom, y descriptions at=edge left},
  width=2.5cm, height=2.5cm, scale only axis,
  enlargelimits=false, axis on top,
  xmin=0, xmax=127, ymin=0, ymax=127,
  xtick={0,63,126}, ytick={0,63,126},
  tick label style={font=\tiny}, title style={font=\footnotesize},
  label style={font=\scriptsize},
  xlabel={$\theta$ bin},
  colormap/viridis, point meta min=0.4, point meta max=0.95,
]
\nextgroupplot[title={$g_1$}, ylabel={Training distribution}]
  \addplot graphics[xmin=0,xmax=127,ymin=0,ymax=127]{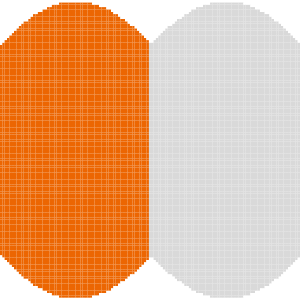};
\nextgroupplot[title={$g_2$}]
  \addplot graphics[xmin=0,xmax=127,ymin=0,ymax=127]{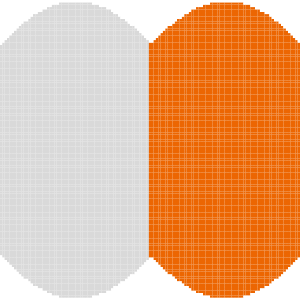};
\nextgroupplot[title={$g_3$}]
  \addplot graphics[xmin=0,xmax=127,ymin=0,ymax=127]{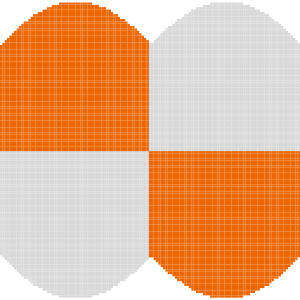};
\nextgroupplot[title={$g_4$},
    colormap={maskmap}{rgb255(0cm)=(217,217,217); rgb255(0.999cm)=(217,217,217);
      rgb255(1cm)=(236,102,2); rgb255(2cm)=(236,102,2)},
    point meta min=-0.5, point meta max=1.5,
    colorbar, colorbar style={width=5pt, at={(1.05,0)}, anchor=south west,
      ytick={0,1}, yticklabels={held-out, sampled},
      tick style={draw=none}, tick label style={font=\tiny}}]
  \addplot graphics[xmin=0,xmax=127,ymin=0,ymax=127]{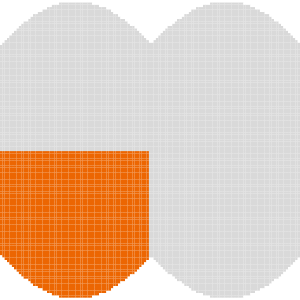};
\nextgroupplot[ylabel={Data-driven}]
  \addplot graphics[xmin=0,xmax=127,ymin=0,ymax=127]{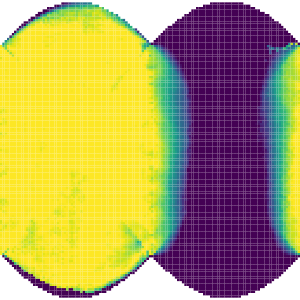};
\nextgroupplot
  \addplot graphics[xmin=0,xmax=127,ymin=0,ymax=127]{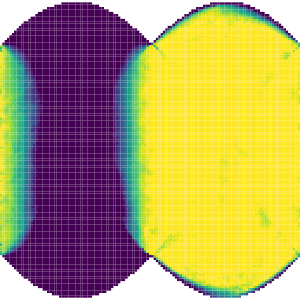};
\nextgroupplot
  \addplot graphics[xmin=0,xmax=127,ymin=0,ymax=127]{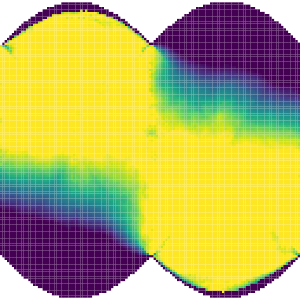};
\nextgroupplot[colorbar, colorbar style={width=5pt, at={(1.05,0)}, anchor=south west,
    ylabel={EA-score}, ylabel style={font=\scriptsize}, ytick={0.4,0.6,0.8},
    tick label style={font=\tiny}}]
  \addplot graphics[xmin=0,xmax=127,ymin=0,ymax=127]{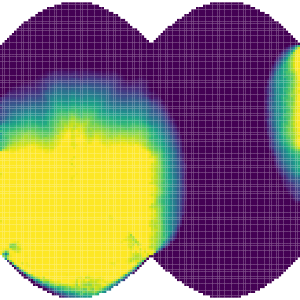};
\nextgroupplot[ylabel={Known-operator}]
  \addplot graphics[xmin=0,xmax=127,ymin=0,ymax=127]{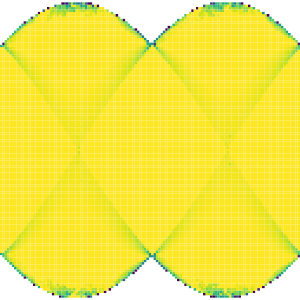};
\nextgroupplot
  \addplot graphics[xmin=0,xmax=127,ymin=0,ymax=127]{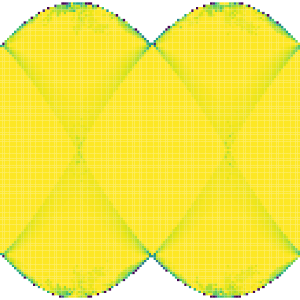};
\nextgroupplot
  \addplot graphics[xmin=0,xmax=127,ymin=0,ymax=127]{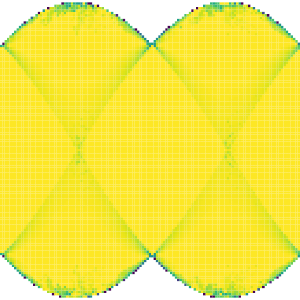};
\nextgroupplot[colorbar, colorbar style={width=5pt, at={(1.05,0)}, anchor=south west,
    ylabel={EA-score}, ylabel style={font=\scriptsize}, ytick={0.4,0.6,0.8},
    tick label style={font=\tiny}}]
  \addplot graphics[xmin=0,xmax=127,ymin=0,ymax=127]{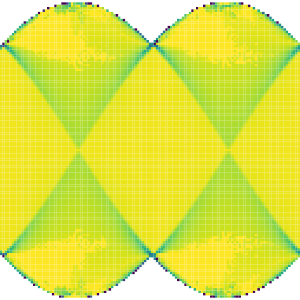};
\end{groupplot}
\end{tikzpicture}
  \caption{Generalisation across training-support settings $g_1$--$g_4$ (columns). Rows: the binary
  training-support region in $(\theta,\rho)$, the data-driven MLP's \glsxtrshort{ea}, and our known-operator
  pipeline's \glsxtrshort{ea} (clean data). The MLP is accurate only where it was trained and collapses on the
  held-out region; the operator stays uniform across the whole line space.}
  \label{fig:generalization}
\end{figure}

Finally we isolate what the known operator itself contributes, comparing our $37.6$K-parameter
pipeline against a far larger data-driven CNN+MLP regressor ($11.3$M parameters, roughly $300\times$ more)
that must learn the recovery from scratch. Both are trained with $\mathcal{L}_{\text{vs}}$, while we restrict
\emph{training} to sub-regions of line space and test on the full space. The four settings
$g_1$--$g_4$ tighten the trained support from a half-space ($g_1,g_2$) through a diagonal split
($g_3$) to a single quadrant ($g_4$), pushing held-out lines progressively farther from anything
seen. 
In-support the two are comparable, out-of-support they part sharply (Figure~\ref{fig:generalization}): the MLP collapses on held-out lines, staying accurate only where it saw data, 
whereas the operator stays almost uniform in- and out-of-support. 
By turning recovery into a fixed geometric computation rather than a learned mapping, the known operator generalises as if the held-out region had been part of training,
and does so with orders of magnitude fewer parameters than the data-driven regressor, which cannot.

	\section{Conclusions}

Geometric prior knowledge in a network is usually placed in the architecture: equivariant layers,
known operators, structured losses. Each of them shapes how evidence is carried, and none of them
ever has to name the object. That obligation falls on the last step. A readout has to return
a single element of the label space itself, so it is the one component that cannot stay agnostic to
how that space is shaped.
An operator that averages coordinates is only ever as correct as the chart it averages in.
What has to change is therefore not the operator but the space it is handed, and that exchange asks
for only three things: an embedding under which the identifications of the label space become
equalities, a linear space to average in, and an inverse in closed form. None of the three introduces
a trainable parameter or a hyperparameter, so whether a readout is well posed is decided by the
construction itself, prior to training and independently of any validation error. Where those three
conditions hold, the seam a chart
introduces has nowhere left to appear, and the \gls{vsmax} is one instance of that.

Lines are not a special case in this respect. Every label space that arises as a quotient by a finite
group action, be it axial orientations, quaternions up to sign, or objects with a discrete symmetry,
admits an embedding of this kind, and the choice of embedding fixes both the
operator and the metric its loss then measures in. The reach of a differentiable argmax is therefore
not bounded by flat charts, but by the embeddings available for the space one wants to read out.

\paragraph{Limitations.} Our evaluation is restricted to single synthetic lines under additive noise.
Real imagery is untested.
This would require a feature extractor that introduces a new uncertain variable in the pipeline.
We intentionally want to isolate this here and dedicate this transition to future work. 
Moreover, it is currently a single-line pipeline only, multi-line settings could further be explored.
Finally, the method is a readout for known-operator line pipelines, and
although the same embed-average-recover recipe should reach other projective label spaces, we have
not demonstrated that.%
	\appendix
\section{Appendix}

\subsection{\texorpdfstring{The squared $L_2$ loss on the Veronese embedding is the squared chordal distance}{The squared L2 loss on the Veronese embedding is the squared chordal distance}}
\label{app:proof}

The training loss is a plain squared $L_2$ distance between the six-dimensional Veronese vectors
$v_2(\mathbf{u})=\operatorname{vech}(\mathbf{u}\mathbf{u}^{\top})$. We show in two steps that, with the
weighting of Section~\ref{sec:veronese}, it equals the squared chordal distance between the lines those
vectors represent. The first step reduces the vector loss to a distance between the outer-product
matrices $\mathbf{u}\mathbf{u}^{\top}$. The vector norm runs over the six distinct entries,
\begin{equation}
  \lVert v_2(\mathbf{u}) - v_2(\mathbf{v})\rVert_2^2 = \sum_{i \leq j} (u_i u_j - v_i v_j)^2 ,
  \label{eq:l2_vector}
\end{equation}
whereas the Frobenius norm runs over all nine matrix entries,
\begin{equation}
  \lVert \mathbf{u}\mathbf{u}^{\top} - \mathbf{v}\mathbf{v}^{\top}\rVert_F^2
  = \sum_{i,j} (u_i u_j - v_i v_j)^2 ,
  \label{eq:frob_matrix}
\end{equation}
so the two differ only in that each off-diagonal pair is counted twice on the right. Scaling the
off-diagonal components by $\sqrt{2}$, which is exactly the weighting
$W=\operatorname{diag}(1,1,1,\sqrt2,\sqrt2,\sqrt2)$ folded into $v_2$ in Section~\ref{sec:veronese},
closes the gap and makes $\lVert W v_2(\mathbf{u}) - W v_2(\mathbf{v})\rVert_2^2 = \lVert
\mathbf{u}\mathbf{u}^{\top} - \mathbf{v}\mathbf{v}^{\top}\rVert_F^2$. Even unweighted the two losses share
the same minimiser and stay monotone in the line angle, so the plain vector loss already trains
correctly.

The second step identifies this Frobenius distance geometrically. For unit vectors
$\mathbf{u},\mathbf{v}\in S^n$, expanding the trace gives
\begin{align}
  \lVert\mathbf{u}\mathbf{u}^{\top} - \mathbf{v}\mathbf{v}^{\top}\rVert_F^2
  &= \operatorname{tr}\!\bigl((\mathbf{u}\mathbf{u}^{\top} - \mathbf{v}\mathbf{v}^{\top})^2\bigr) \notag \\
  &= \operatorname{tr}(\mathbf{u}\mathbf{u}^{\top}\mathbf{u}\mathbf{u}^{\top})
     - 2\operatorname{tr}(\mathbf{u}\mathbf{u}^{\top}\mathbf{v}\mathbf{v}^{\top})
     + \operatorname{tr}(\mathbf{v}\mathbf{v}^{\top}\mathbf{v}\mathbf{v}^{\top}) \notag \\
  &= \lVert\mathbf{u}\rVert^4 - 2(\mathbf{u}^{\top}\mathbf{v})^2 + \lVert\mathbf{v}\rVert^4 \notag \\
  &= 2 - 2(\mathbf{u}^{\top}\mathbf{v})^2 \notag \\
  &= 2\bigl(1 - \cos^2\alpha\bigr) \notag \\
  &= 2\sin^2\alpha ,
  \label{eq:proof_chordal}
\end{align}
where $\alpha=\arccos\lvert\mathbf{u}^{\top}\mathbf{v}\rvert\in[0,\tfrac{\pi}{2}]$ is the angle between
the projective points $[\mathbf{u}],[\mathbf{v}]\in\mathbb{RP}^n$. The right-hand side is their squared
chordal distance, the squared Euclidean distance between the projector embeddings
\citep{rotation_averaging}, and it is invariant to the signs of $\mathbf{u}$ and $\mathbf{v}$, hence
well defined on undirected lines ($n=2$). Chaining the two steps, the squared $L_2$ loss on the
weighted Veronese embedding is exactly the squared chordal distance between the corresponding lines.

\subsection{Möbius padding for the single cover}
\label{app:mobius}

The single-cover accumulator $A\in\mathbb{R}^{T\times R}$ indexes orientation by
$t\in\{0,\dots,T-1\}$ over $\theta\in[0,\pi)$ and signed offset by $r\in\{0,\dots,R-1\}$ symmetric
about the centre bin, so that $r\mapsto R-1-r$ realises $\rho\mapsto-\rho$ (with $R=127$ odd the
centre bin is exact). The seam identification $(\theta,\rho)\sim(\theta+\pi,-\rho)$ fixes the
topologically correct padding: a virtual row past either angular boundary is the opposite boundary
with $\rho$ reflected,
\begin{equation}
  A[-1-k,\,r] \;=\; A[\,T-1-k,\ R-1-r\,],\qquad
  A[\,T+k,\,r\,] \;=\; A[\,k,\ R-1-r\,],\qquad k\ge0 .
\end{equation}
This is circular padding in $\theta$ composed with a reflection in $\rho$---a glide-reflection. It
splices in the exact heatmap values of the continuing lines and is the most faithful single-cover
padding available. It cannot restore equivariance: a plane convolution is equivariant to
translations but not to the orientation-reversing glide-reflection the Möbius seam requires, so an
oriented kernel meets a $\rho$-reflected continuation across the boundary (Section~\ref{sec:veronese}).
The double cover replaces this glide-reflection with a pure translation, for which ordinary circular
padding is exact.

\subsection{RHT: a multi-line detector applied to single-line regression}
\label{app:rht}

The \gls{rht} is a learned Hough baseline from the multi-line detection
literature. A network regresses a heatmap over the $(\theta,\rho)$ accumulator whose peaks mark the
lines, trained to match Gaussian targets of width $\sigma_t$; the lines are then recovered by a
separate, non-differentiable peak-extraction step. The method is built to detect an unknown number
of lines rather than to regress a single line end-to-end, and its readout cannot be trained through.
We evaluate it here as a Hough-based reference. Because the target width $\sigma_t$ is a free
hyperparameter uncoupled from the line error, we sweep it widely; Figure~\ref{fig:rht-sweep} reports
global \gls{ea} on the full test set for $\sigma_t\in\{0.5,1.1,1.5,2.0,3.0\}$. On clean data a wider target
helps---\gls{ea} rises from $0.79$ at $\sigma_t{=}0.5$ to $0.89$ at $\sigma_t{=}3.0$---but under even mild
noise every width collapses to $\text{EA}\approx0.1$ by $\sigma{=}0.3$, and no configuration we tried
recovers. We conclude the recipe does not transfer to single-line regression, and adopt its strongest
member ($\sigma_t{=}3.0$) as the RHT reference in Figure~\ref{fig:endtoend}.

\begin{figure}[H]
  \centering
\begin{tikzpicture}
\begin{axis}[paper, width=\linewidth, height=5cm,
  xlabel={test noise $\sigma$},
  ylabel={EA-score},
  xmin=0, xmax=0.8, ymin=0, ymax=1.02,
  xtick={0,0.1,0.2,0.3,0.4,0.5,0.6,0.7,0.8},
  toplegend, legend columns=-1,
  cycle list={
    {colBase, mark=*, mark size=1pt},
    {colMlp, mark=square*, mark size=1pt},
    {colDsnt, mark=triangle*, mark size=1.3pt},
    {colHough, mark=diamond*, mark size=1.3pt},
    {colOursB, mark=pentagon*, mark size=1.3pt}},
]
  \addplot+[line width=1.0pt] table[x=sigma, y=rht_s05]{figures/dat/ea_rht_sweep.dat};
  \addlegendentry{$\sigma_t{=}0.5$}
  \addplot+[line width=1.0pt] table[x=sigma, y=rht_s11]{figures/dat/ea_rht_sweep.dat};
  \addlegendentry{$\sigma_t{=}1.1$}
  \addplot+[line width=1.0pt] table[x=sigma, y=rht_s15]{figures/dat/ea_rht_sweep.dat};
  \addlegendentry{$\sigma_t{=}1.5$}
  \addplot+[line width=1.0pt] table[x=sigma, y=rht_s20]{figures/dat/ea_rht_sweep.dat};
  \addlegendentry{$\sigma_t{=}2.0$}
  \addplot+[line width=1.1pt] table[x=sigma, y=rht_s30]{figures/dat/ea_rht_sweep.dat};
  \addlegendentry{$\sigma_t{=}3.0$}
\end{axis}
\end{tikzpicture}
  \caption{RHT global \glsxtrshort{ea} on the full test set vs.\ test noise $\sigma$, per target-heatmap width
  $\sigma_t$. A wider target helps when clean but every width collapses once noise appears---no
  target width recovers.}
  \label{fig:rht-sweep}
\end{figure}

\subsection{Training setup and baseline configurations}
\label{app:setup}

\begin{table}[H]
  \centering
  \caption{Optimisation, batching and baseline configurations. The optimisation and batching blocks
  are identical across every run reported in the paper.}
  \label{tab:setup}
  \begin{tabular}{ll}
    \toprule
    \multicolumn{2}{l}{\emph{Optimisation}} \\
    \midrule
    Optimiser              & AdamW \\
    Learning rate          & $10^{-4}$ \\
    Weight decay           & $10^{-4}$ \\
    Betas                  & $(0.9,\,0.999)$ \\
    Scheduler              & ReduceLROnPlateau (factor $0.1$, patience $10$) \\
    Early stopping         & on val.\ loss (patience $20$, min.\ delta $10^{-4}$) \\
    Max.\ epochs           & $1000$ \\
    Model selection        & best validation loss \\
    \midrule
    \multicolumn{2}{l}{\emph{Batching}} \\
    \midrule
    Batch size             & $64$ (train / validation / test) \\
    Seed                   & $42$, deterministic \\
    \midrule
    \multicolumn{2}{l}{\emph{Baseline}} \\
    \midrule
    CNN+MLP                & ResNet-18 (ImageNet-initialised, fine-tuned) \\
                           & $+$ MLP head with $256$ hidden units, $11.3$M parameters \\
    \bottomrule
  \end{tabular}
\end{table}

\begin{tikzpicture}[>=stealth, font=\small,
  block/.style={rounded corners=3pt, draw=gray!70, fill=gray!8, font=\footnotesize,
                minimum width=1.9cm, minimum height=0.96cm, align=center},
  ours/.style={rounded corners=3pt, draw=healthy_orange, line width=0.9pt, fill=healthy_orange!10,
               font=\footnotesize, minimum width=1.9cm, minimum height=0.96cm, align=center},
  cnn/.style={rounded corners=2pt, draw=gray!70, fill=gray!8,
              minimum width=0.7cm, minimum height=0.4cm, inner sep=1pt, font=\scriptsize},
  grad/.style={gray!80, opacity=0.4, line width=0.6pt, ->},
  wire/.style={gray!60, line width=0.3pt}]

\begin{scope}[yshift=2.4cm]
  \node[font=\small\bfseries, anchor=east] at (-0.35,0) {(a)};

  \begin{scope}
    \clip (0,-0.75) rectangle (1.5,0.75);
    \fill[black] (0,-0.75) rectangle (1.5,0.75);
    \draw[white, line width=0.9pt] (0.15,-0.9) -- (1.35,0.9);
  \end{scope}
  \draw[gray!60] (0,-0.75) rectangle (1.5,0.75);

  \draw[->]   (1.55,-0.10) -- (2.00,-0.10);
  \draw[grad] (2.00, 0.10) -- (1.55, 0.10);

  \node[cnn, minimum width=1.0cm, minimum height=0.9cm] at (2.55,0) {CNN};

  \foreach \y in {0.45,0.15,-0.15,-0.45}{%
    \foreach \z in {0.2,-0.2}{\draw[wire] (3.425,\y) -- (4.425,\z);}}
  \foreach \y in {0.45,0.15,-0.15,-0.45}{\fill[gray!55] (3.425,\y) circle (1.6pt);}
  \foreach \z in {0.2,-0.2}{\fill[gray!55] (4.425,\z) circle (1.6pt);}
  \node[anchor=west, font=\footnotesize] at (4.575,0.2) {$p_1$};
  \node[anchor=west, font=\footnotesize] at (4.575,-0.2) {$p_2$};
  \draw[gray!70, line width=0.5pt, decorate, decoration={brace, amplitude=4pt}]
    (5.025,0.33) -- (5.025,-0.33) node[midway, right=5pt, font=\footnotesize, text=black]{$\vec{p}$};

\end{scope}

\begin{scope}
  \node[font=\small\bfseries, anchor=east] at (-0.35,0) {(b)};

  \begin{scope}
    \clip (0,-0.75) rectangle (1.5,0.75);
    \fill[black] (0,-0.75) rectangle (1.5,0.75);
    \draw[white, line width=0.9pt] (0.15,-0.9) -- (1.35,0.9);
  \end{scope}
  \draw[gray!60] (0,-0.75) rectangle (1.5,0.75);

  \draw[->]   (1.55,-0.10) -- (2.00,-0.10);
  \draw[grad] (2.00, 0.10) -- (1.55, 0.10);

  \node[block] at (3.00,0) {HT-Layer};

  \draw[->]   (4.00,-0.10) -- (4.45,-0.10);
  \draw[grad] (4.45, 0.10) -- (4.00, 0.10);

  \begin{scope}[shift={(4.50,-0.48)}]
    \begin{scope}
      \clip (0,0) rectangle (0.96,0.96);
      \fill[black] (0,0) rectangle (0.96,0.96);
      \foreach \s in {-0.8,-0.7,...,0.8}{%
        \draw[white, opacity=0.3, line width=0.2pt]
          plot[domain=0:180, samples=80, smooth]
            ({0.96*\x/180}, {0.48+0.48*(0.2*cos(\x-115)+\s*sin(\x-115))});}
      \fill[white] (0.613,0.576) circle (1.3pt);
    \end{scope}
    \draw[gray!60] (0,0) rectangle (0.96,0.96);
  \end{scope}

  \draw[->]   (5.51,-0.10) -- (5.96,-0.10);
  \draw[grad] (5.96, 0.10) -- (5.51, 0.10);

  \node[cnn] at (6.36,0) {CNN};

  \draw[->]   (6.76,-0.10) -- (7.21,-0.10);
  \draw[grad] (7.21, 0.10) -- (6.76, 0.10);

  \begin{scope}[shift={(7.26,-0.48)}]
    \fill[black] (0,0) rectangle (0.96,0.96);
    \fill[white] (0.613,0.576) circle (1.3pt);
    \draw[gray!60] (0,0) rectangle (0.96,0.96);
  \end{scope}

  \draw[->]   (8.27,-0.10) -- (8.72,-0.10);
  \draw[grad] (8.72, 0.10) -- (8.27, 0.10);

  \node[ours] at (9.72,0) {Readout};

  \draw[->]   (10.72,-0.10) -- (11.17,-0.10);
  \draw[grad] (11.17, 0.10) -- (10.72, 0.10);
  \node[anchor=west] at (11.22,0) {$\vec{p}$};
\end{scope}

\end{tikzpicture}%
\begin{tikzpicture}[>=stealth, font=\footnotesize]

\begin{scope}
  \draw[gray!60] (0,0) rectangle (3.4,3.4);
  \draw[healthy_orange, line width=1pt] (2.30,0) -- (2.50,3.4);
  \draw[siemens_petrol, line width=1pt, densely dashed] (2.50,0) -- (2.30,3.4);
  \node[healthy_orange, anchor=west] at (2.60,0.80) {$\ell_1$};
  \node[siemens_petrol,  anchor=east] at (2.20,2.60) {$\ell_2$};
  \node[align=center] at (1.7,-0.95) {image space\\[-2pt]\scriptsize two nearly identical lines};
\end{scope}

\begin{scope}[xshift=5.2cm]
  \draw[gray!60] (0,0) rectangle (3.4,3.4);
  \draw[gray!35, dashed] (0,1.7) -- (3.4,1.7);

  \fill[healthy_orange] (0.16,2.66) circle (2pt);
  \node[healthy_orange, anchor=west] at (0.30,2.66) {$\ell_1$};
  \fill[siemens_petrol] (3.24,0.74) circle (2pt);
  \node[siemens_petrol, anchor=east] at (3.10,0.74) {$\ell_2$};

  \draw[healthy_orange, ->, line width=1.4pt] (0,0.35) -- (0,3.05);
  \draw[siemens_petrol,  ->, line width=1.4pt] (3.4,3.05) -- (3.4,0.35);

  \node[anchor=north] at (0,-0.05) {\scriptsize $0$};
  \node[anchor=north] at (3.4,-0.05) {\scriptsize $\pi$};
  \node[anchor=north] at (1.7,-0.05) {$\theta$};
  \node[rotate=90] at (-0.38,1.7) {$\rho$};
  \node[align=center] at (1.7,-0.95) {Hough space\\[-2pt]\scriptsize opposite corners, glued with a flip};
\end{scope}

\end{tikzpicture}%
\begin{tikzpicture}
\begin{axis}[paper, width=\linewidth, height=5cm,
  xlabel={orientation accumulator bin $\theta$ \;(seam at $0$ and $126$)},
  ylabel={EA-score},
  xmin=0, xmax=126, ymin=-0.02, ymax=1.05,
  xtick={0,21,42,63,84,105,126},
  toplegend,
]
  \addplot[name path=dsntlo, draw=none, forget plot] table[x=tbin, y=dsntve_lo]{figures/dat/ea_theta_exp1.dat};
  \addplot[name path=dsnthi, draw=none, forget plot] table[x=tbin, y=dsntve_hi]{figures/dat/ea_theta_exp1.dat};
  \addplot[draw=none, fill=colDsnt, fill opacity=0.16, forget plot] fill between[of=dsntlo and dsnthi];
  \addplot[colDsnt, line width=1.0pt] table[x=tbin, y=dsntve]{figures/dat/ea_theta_exp1.dat};
  \addlegendentry{soft-argmax}
  \addplot[name path=ourslo, draw=none, forget plot] table[x=tbin, y=vsve2_lo]{figures/dat/ea_theta_exp1.dat};
  \addplot[name path=ourshi, draw=none, forget plot] table[x=tbin, y=vsve2_hi]{figures/dat/ea_theta_exp1.dat};
  \addplot[draw=none, fill=colOursB, fill opacity=0.16, forget plot] fill between[of=ourslo and ourshi];
  \addplot[colOursB, line width=1.2pt] table[x=tbin, y=vsve2]{figures/dat/ea_theta_exp1.dat};
  \addlegendentry{Veronese soft-argmax (ours)}
\end{axis}
\end{tikzpicture}%
\begin{tikzpicture}
\begin{groupplot}[
  group style={group size=2 by 1, horizontal sep=0.9cm},
  width=3.2cm, height=3.2cm, scale only axis,
  enlargelimits=false, axis on top,
  xmin=0, xmax=127, ymin=0, ymax=127,
  xtick={0,63,126}, ytick={0,63,126},
  tick label style={font=\scriptsize}, title style={font=\footnotesize},
  xlabel={$\theta$ bin}, xlabel style={font=\scriptsize},
  colormap/viridis, point meta min=0.4, point meta max=0.95,
]
\nextgroupplot[title={soft-argmax}, ylabel={$\rho$ bin}, ylabel style={font=\scriptsize}]
  \addplot graphics[xmin=0, xmax=127, ymin=0, ymax=127]{figures/plots/ea_dsntve.pdf};
\nextgroupplot[title={VSmax},
    colorbar, colorbar style={width=5pt, ylabel={EA-score}, ylabel style={font=\scriptsize},
    ytick={0.4,0.6,0.8}, tick label style={font=\scriptsize}}]
  \addplot graphics[xmin=0, xmax=127, ymin=0, ymax=127]{figures/plots/ea_vsve2.pdf};
\end{groupplot}
\end{tikzpicture}%
\begin{tikzpicture}
\begin{groupplot}[
  group style={group size=3 by 1, horizontal sep=1.2cm},
  paper, width=0.30\linewidth, height=4.4cm,
  xmin=0, xmax=0.8, ymin=0, ymax=1.02,
  xtick={0,0.2,0.4,0.6,0.8},
  xlabel={test noise $\sigma$},
  title style={font=\footnotesize},
]
\nextgroupplot[title={MLP}, ylabel={interior EA-score}]
  \addplot[healthy_orange, line width=1.1pt, mark=*, mark size=1pt] table[x=sigma, y=mlpve]{figures/dat/ea_noise_interior.dat};
  \addplot[siemens_petrol, line width=1.0pt, densely dashed, mark=triangle*, mark size=1.3pt] table[x=sigma, y=mlppolar]{figures/dat/ea_noise_interior.dat};
\nextgroupplot[title={soft-argmax}]
  \addplot[healthy_orange, line width=1.1pt, mark=*, mark size=1pt] table[x=sigma, y=dsntve]{figures/dat/ea_noise_interior.dat};
  \addplot[siemens_petrol, line width=1.0pt, densely dashed, mark=triangle*, mark size=1.3pt] table[x=sigma, y=dsntpolar]{figures/dat/ea_noise_interior.dat};
\nextgroupplot[title={VSmax},
  legend pos=south west, legend style={draw=none, fill=none, font=\scriptsize}]
  \addplot[healthy_orange, line width=1.1pt, mark=*, mark size=1pt] table[x=sigma, y=vsve2]{figures/dat/ea_noise_interior.dat};
  \addlegendentry{$\mathcal{L}_{\text{vs}}$}
  \addplot[siemens_petrol, line width=1.0pt, densely dashed, mark=triangle*, mark size=1.3pt] table[x=sigma, y=vspolar2]{figures/dat/ea_noise_interior.dat};
  \addlegendentry{$\mathcal{L}_{\text{polar}}$}
\end{groupplot}
\end{tikzpicture}%
\begin{tikzpicture}
\begin{groupplot}[
  group style={group size=4 by 3, horizontal sep=0.45cm, vertical sep=0.6cm,
    x descriptions at=edge bottom, y descriptions at=edge left},
  width=2.5cm, height=2.5cm, scale only axis,
  enlargelimits=false, axis on top,
  xmin=0, xmax=127, ymin=0, ymax=127,
  xtick={0,63,126}, ytick={0,63,126},
  tick label style={font=\tiny}, title style={font=\footnotesize},
  label style={font=\scriptsize},
  xlabel={$\theta$ bin},
  colormap/viridis, point meta min=0.4, point meta max=0.95,
]
\nextgroupplot[title={$g_1$}, ylabel={Training distribution}]
  \addplot graphics[xmin=0,xmax=127,ymin=0,ymax=127]{figures/plots/mask_g1.pdf};
\nextgroupplot[title={$g_2$}]
  \addplot graphics[xmin=0,xmax=127,ymin=0,ymax=127]{figures/plots/mask_g2.pdf};
\nextgroupplot[title={$g_3$}]
  \addplot graphics[xmin=0,xmax=127,ymin=0,ymax=127]{figures/plots/mask_g3.pdf};
\nextgroupplot[title={$g_4$},
    colormap={maskmap}{rgb255(0cm)=(217,217,217); rgb255(0.999cm)=(217,217,217);
      rgb255(1cm)=(236,102,2); rgb255(2cm)=(236,102,2)},
    point meta min=-0.5, point meta max=1.5,
    colorbar, colorbar style={width=5pt, at={(1.05,0)}, anchor=south west,
      ytick={0,1}, yticklabels={held-out, sampled},
      tick style={draw=none}, tick label style={font=\tiny}}]
  \addplot graphics[xmin=0,xmax=127,ymin=0,ymax=127]{figures/plots/mask_g4.pdf};
\nextgroupplot[ylabel={Data-driven}]
  \addplot graphics[xmin=0,xmax=127,ymin=0,ymax=127]{figures/plots/ea_mlp_g1.pdf};
\nextgroupplot
  \addplot graphics[xmin=0,xmax=127,ymin=0,ymax=127]{figures/plots/ea_mlp_g2.pdf};
\nextgroupplot
  \addplot graphics[xmin=0,xmax=127,ymin=0,ymax=127]{figures/plots/ea_mlp_g3.pdf};
\nextgroupplot[colorbar, colorbar style={width=5pt, at={(1.05,0)}, anchor=south west,
    ylabel={EA-score}, ylabel style={font=\scriptsize}, ytick={0.4,0.6,0.8},
    tick label style={font=\tiny}}]
  \addplot graphics[xmin=0,xmax=127,ymin=0,ymax=127]{figures/plots/ea_mlp_g4.pdf};
\nextgroupplot[ylabel={Known-operator}]
  \addplot graphics[xmin=0,xmax=127,ymin=0,ymax=127]{figures/plots/ea_op_g1.pdf};
\nextgroupplot
  \addplot graphics[xmin=0,xmax=127,ymin=0,ymax=127]{figures/plots/ea_op_g2.pdf};
\nextgroupplot
  \addplot graphics[xmin=0,xmax=127,ymin=0,ymax=127]{figures/plots/ea_op_g3.pdf};
\nextgroupplot[colorbar, colorbar style={width=5pt, at={(1.05,0)}, anchor=south west,
    ylabel={EA-score}, ylabel style={font=\scriptsize}, ytick={0.4,0.6,0.8},
    tick label style={font=\tiny}}]
  \addplot graphics[xmin=0,xmax=127,ymin=0,ymax=127]{figures/plots/ea_op_g4.pdf};
\end{groupplot}
\end{tikzpicture}%
\begin{tikzpicture}
\begin{axis}[paper, width=\linewidth, height=5cm,
  xlabel={test noise $\sigma$},
  ylabel={EA-score},
  xmin=0, xmax=0.8, ymin=0, ymax=1.02,
  xtick={0,0.1,0.2,0.3,0.4,0.5,0.6,0.7,0.8},
  toplegend,
]
  \addplot[healthy_orange!55, line width=1.0pt, mark=square*, mark size=1pt] table[x=sigma, y=rht]{figures/dat/ea_noise_full.dat};
  \addlegendentry{RHT}
  \addplot[healthy_orange, line width=1.0pt, densely dashed, mark=diamond*, mark size=1.3pt] table[x=sigma, y=classicalhough]{figures/dat/ea_noise_full.dat};
  \addlegendentry{HT+argmax}
  \addplot[healthy_orange!70!black, line width=1.2pt, mark=*, mark size=1pt] table[x=sigma, y=vsve2]{figures/dat/ea_noise_full.dat};
  \addlegendentry{VSmax}
\end{axis}
\end{tikzpicture}%
\begin{tikzpicture}
\begin{axis}[paper, width=\linewidth, height=5cm,
  xlabel={test noise $\sigma$},
  ylabel={EA-score},
  xmin=0, xmax=0.8, ymin=0, ymax=1.02,
  xtick={0,0.1,0.2,0.3,0.4,0.5,0.6,0.7,0.8},
  toplegend, legend columns=-1,
  cycle list={
    {colBase, mark=*, mark size=1pt},
    {colMlp, mark=square*, mark size=1pt},
    {colDsnt, mark=triangle*, mark size=1.3pt},
    {colHough, mark=diamond*, mark size=1.3pt},
    {colOursB, mark=pentagon*, mark size=1.3pt}},
]
  \addplot+[line width=1.0pt] table[x=sigma, y=rht_s05]{figures/dat/ea_rht_sweep.dat};
  \addlegendentry{$\sigma_t{=}0.5$}
  \addplot+[line width=1.0pt] table[x=sigma, y=rht_s11]{figures/dat/ea_rht_sweep.dat};
  \addlegendentry{$\sigma_t{=}1.1$}
  \addplot+[line width=1.0pt] table[x=sigma, y=rht_s15]{figures/dat/ea_rht_sweep.dat};
  \addlegendentry{$\sigma_t{=}1.5$}
  \addplot+[line width=1.0pt] table[x=sigma, y=rht_s20]{figures/dat/ea_rht_sweep.dat};
  \addlegendentry{$\sigma_t{=}2.0$}
  \addplot+[line width=1.1pt] table[x=sigma, y=rht_s30]{figures/dat/ea_rht_sweep.dat};
  \addlegendentry{$\sigma_t{=}3.0$}
\end{axis}
\end{tikzpicture}%
\newcommand{\runlabelmain}[1]{%
  \ifcase#1
    \lblmlppolar\or    
    \lblmlpve\or       
    \lbldsntpolar\or   
    \lbldsntve\or      
    \lblvspolartwo\or  
    \lblvsvetwo\fi     
}

\pgfplotstabletypeset[
  col sep=space,
  every head row/.style={before row=\toprule, after row=\midrule},
  every last row/.style={after row=\bottomrule},
  every row no. 2/.style={before row=\midrule},
  every row no. 4/.style={before row=\midrule},
  columns={run,ea_all,ea_seam,ea_interior},
  columns/run/.style={
    column name={Method}, column type={l},
    assign cell content/.code={\pgfkeyssetvalue{/pgfplots/table/@cell content}{\runlabelmain{##1}}},
  },
  columns/ea_all/.style={column name={EA (all)},           string type, column type={c}},
  columns/ea_seam/.style={column name={EA (seam)},         string type, column type={c}},
  columns/ea_interior/.style={column name={EA (interior)}, string type, column type={c}},
]{figures/dat/ea_methods_main.dat}
\newcommand{\runlabelbound}[1]{%
  \ifcase#1 \lbldsntve\or \lblvsvetwo\fi
}

\pgfplotstabletypeset[
  col sep=space,
  every head row/.style={before row=\toprule, after row=\midrule},
  every last row/.style={after row=\bottomrule},
  columns={run,ea_seam,ea_interior,ea_all},
  columns/run/.style={
    column name={Readout}, column type={l},
    assign cell content/.code={\pgfkeyssetvalue{/pgfplots/table/@cell content}{\runlabelbound{##1}}},
  },
  columns/ea_seam/.style={column name={EA (seam)},         fixed, fixed zerofill, precision=3},
  columns/ea_interior/.style={column name={EA (interior)}, fixed, fixed zerofill, precision=3},
  columns/ea_all/.style={column name={EA (all)},           fixed, fixed zerofill, precision=3},
]{figures/dat/ea_boundary_exp1.dat}
\pgfplotstabletypeset[
  col sep=space,
  every head row/.style={before row={\toprule
    & \multicolumn{2}{c}{MLP regression} & \multicolumn{2}{c}{Hough operator (ours)}\\
    \cmidrule(lr){2-3}\cmidrule(lr){4-5}}, after row=\midrule},
  every last row/.style={after row=\bottomrule},
  columns={g,mlp_in,mlp_out,op_in,op_out},
  columns/g/.style={column name={Setting}, column type={l},
    assign cell content/.code={\pgfkeyssetvalue{/pgfplots/table/@cell content}{$g_{##1}$}}},
  columns/mlp_in/.style={column name={in-support},  string type, column type={c}},
  columns/mlp_out/.style={column name={held-out},   string type, column type={c}},
  columns/op_in/.style={column name={in-support},   string type, column type={c}},
  columns/op_out/.style={column name={held-out},    string type, column type={c}},
]{figures/dat/gen_generalization.dat}
	\input{figures/dat/ea_boundary_exp1.dat}%
	\input{figures/dat/ea_methods_main.dat}%
	\input{figures/dat/ea_noise_full.dat}%
	\input{figures/dat/ea_noise_interior.dat}%
	\input{figures/dat/ea_rht_sweep.dat}%
	\input{figures/dat/ea_theta_exp1.dat}%
	\input{figures/dat/gen_generalization.dat}%
	\includegraphics{figures/plots/ea_dsntve.pdf}%
	\includegraphics{figures/plots/ea_vsve2.pdf}%
	\includegraphics{figures/plots/mask_g1.pdf}%
	\includegraphics{figures/plots/mask_g2.pdf}%
	\includegraphics{figures/plots/mask_g3.pdf}%
	\includegraphics{figures/plots/mask_g4.pdf}%
	\includegraphics{figures/plots/ea_mlp_g1.pdf}%
	\includegraphics{figures/plots/ea_mlp_g2.pdf}%
	\includegraphics{figures/plots/ea_mlp_g3.pdf}%
	\includegraphics{figures/plots/ea_mlp_g4.pdf}%
	\includegraphics{figures/plots/ea_op_g1.pdf}%
	\includegraphics{figures/plots/ea_op_g2.pdf}%
	\includegraphics{figures/plots/ea_op_g3.pdf}%
	\includegraphics{figures/plots/ea_op_g4.pdf}%
}

\definecolor{healthy_orange}{HTML}{EC6602}
\definecolor{siemens_petrol}{HTML}{009999}

\title{\centering Soft-Argmax for the Projective Plane \\ via the Veronese Embedding}

\author{
\parbox[t]{\textwidth}{\centering
\normalfont
Benjamin El-Zein\textsuperscript{1,2}, Dominik Eckert\textsuperscript{1}, Paul Zech\textsuperscript{1,2},
Christopher Syben\textsuperscript{1},\\
Bernhard Geiger\textsuperscript{1}, Steffen Kappler\textsuperscript{1}, Sebastian Stober\textsuperscript{2}\\[0.7em]
\small
\textsuperscript{1}Siemens Healthineers AG, X-ray Products, Forchheim, Germany\\
\textsuperscript{2}Artificial Intelligence Lab, Otto-von-Guericke-University, Magdeburg, Germany}
}

\iclrfinalcopy 

\begin{document}
\begin{filecontents*}{figures/dat/ea_methods_main.dat}
run ea_all ea_seam ea_interior
0 0.719$\pm$0.268 0.561$\pm$0.297 0.730$\pm$0.263
1 0.763$\pm$0.294 0.720$\pm$0.293 0.766$\pm$0.294
2 0.659$\pm$0.314 0.627$\pm$0.351 0.662$\pm$0.311
3 0.717$\pm$0.298 0.692$\pm$0.339 0.719$\pm$0.295
4 0.692$\pm$0.295 0.806$\pm$0.237 0.684$\pm$0.297
5 0.878$\pm$0.233 0.918$\pm$0.177 0.875$\pm$0.237
\end{filecontents*}

\begin{filecontents*}{figures/dat/ea_boundary_exp1.dat}
run ea_all ea_seam ea_interior
0 0.717099 0.691986 0.718855
1 0.878249 0.918245 0.875452
\end{filecontents*}

\begin{filecontents*}{figures/dat/gen_generalization.dat}
g mlp_in mlp_out op_in op_out
1 0.934$\pm$0.147 0.325$\pm$0.341 0.949$\pm$0.058 0.949$\pm$0.058
2 0.945$\pm$0.116 0.289$\pm$0.330 0.948$\pm$0.058 0.947$\pm$0.059
3 0.931$\pm$0.204 0.515$\pm$0.331 0.948$\pm$0.054 0.948$\pm$0.055
4 0.956$\pm$0.085 0.296$\pm$0.315 0.914$\pm$0.060 0.914$\pm$0.063
\end{filecontents*}

\maketitle
\lhead{} 
\renewcommand{\headrulewidth}{0pt} 

\begin{abstract}
From horizon detection to fibre structures in X-ray imaging, many vision tasks recover lines via peak detection in Hough space $H=S^1\times\mathbb{R}$, the domain of orientation-offset pairs $(\theta,\rho)$.
Differentiable pipelines extract coordinates via \emph{soft-argmax}, a probability-weighted average that is only meaningful in a globally linear space.
However, $(\theta,\rho)$ and $(\theta+\pi,-\rho)$ describe the same undirected line, so $H$ double-covers the space of undirected lines $H/\mathbb{Z}_2$: a Möbius strip, obtained by identifying each pair under $\mathbb{Z}_2$ action.
Soft-argmax operates on the cover $H$, but since $H/\mathbb{Z}_2$ admits no linear structure, it tears geometrically adjacent lines apart. 
Thus we need a $\mathbb{Z}_2$-invariant embedding of lines into a linear space, on which soft-argmax is well-defined.
We achieve this by parametrising lines via unit-norm homogeneous vectors $\ell=(1+\rho^2)^{-1/2}(\cos\theta,\sin\theta,-\rho)^{\top}\in\mathbb{R}^3$ and applying the Veronese map $v_2(\ell)=\ell\ell^{\top}$ that satisfies $v_2(\ell)=v_2(-\ell)$.
This descends continuously to an embedding of the quotient $H/\mathbb{Z}_2$ into the linear space $\mathrm{Sym}^2(\mathbb{R}^3)$, where the antipodal ambiguity vanishes.
Line extraction becomes a barycentre in $\mathrm{Sym}^2(\mathbb{R}^3)$, projected back via its leading eigenvector.
We validate our \emph{Veronese soft-argmax} in a Hough transform-based network across all resolvable lines, confirming uniform and seam-free recovery.
We further derive that the $L_2$-loss on isometrically weighted Veronese embeddings equals the squared chordal distance between lines in projective space, enabling a geometrically precise training objective.
\end{abstract}

\vspace{1em}
\noindent{\footnotesize\textbf{Disclaimer:}\quad The presented methods in this paper are not
commercially available and their future availability cannot be guaranteed.\par}

\clearpage
\bibliography{references/zotero}
\bibliographystyle{references/iclr}

\clearpage

\end{document}